\documentclass[twocolumn]{rps-esrl2020}

\def\papername{\jobname}

\usepackage{algorithm}
\usepackage{algorithmic}
\usepackage{amsmath}
\usepackage{enumitem}

\begin{document}

%\markboth{First Author and Second Author}{Instructions for Preparing Paper for ESREL 2020 PSAM 15}

%%%%%%%%%%%%%%%%%%%%%%%%% Plase keep this command for single column for abstract section.
\twocolumn[
%%%%%%%%%%%%%%%%%%%%%%%%%

\title{Neural Architecture Search For Fault Diagnosis}

\author{Xudong Li}
\address{National Space Science Center.CAS, University of Chinese Academy of Sciences, Beijing,China. \email{lixudong16@mails.ucas.edu.cn}}

\author{Yang Hu}
\address{Science and Technology on Complex Aviation System Simulation Laboratory, 9236 mailbox, Beijing, China. \email{yang.hu@polimi.it}}

\author{Jianhua Zheng}
\address{National Space Science Center.CAS, University of Chinese Academy of Sciences, Beijing,China. \email{zhengjianhua@nssc.ac.cn}}

\author{Mingtao Li}
\address{National Space Science Center.CAS, University of Chinese Academy of Sciences, Beijing,China. \email{limingtao@nssc.ac.cn}}

\begin{abstract} 
Data-driven methods has made great progress in fault diagnosis, especially deep learning method.  Deep learning is suitable for processing big data, and has a strong feature extraction ability to realize end-to-end fault diagnosis systems. Due to the complexity and variability of experimental data, some deep learning models are designed to be complex.  However, designing neural network architecture requires rich professional knowledge and debugging experience, and a lot of experiments are needed to screen models and hyperparameters, increasing the difficulty of developing deep learning models.  Fortunately, neural architecture search (NAS) is developing rapidly, and is becoming one of the next directions for deep learning.  Given a search space,  NAS can automatically search for the optimal network architecture.  In this paper, we proposed a NAS method for fault diagnosis using reinforcement learning.  A recurrent neural network is used as an agent to generate the network architecture. The accuracy of the generated network on the validation dataset is fed back to the agent as a reward, and the parameters of the agent are updated through the strategy gradient algorithm.  In order to speed up the search process, parameters sharing method is adopted in this paper.  We use PHM 2009 Data Challeng gearbox dataset to prove the effectiveness of propsed method, and obtain state-of-the-art results compared with other artificial designed network structures.  To author’s best knowledge, it’s the first time that NAS has been applied in fault diagnosis.
\end{abstract}


%%%%%%%%%%%%%%%%%%%%%%%%% Please keep this closing bracket to complete the single column format for abstract.
]
%%%%%%%%%%%%%%%%%%%%%%%%%

\section{Introduction}
In recent years, deep learning has been successfully applied in fault diagnosis, and it has become a new research hotspot in data-driven methods \cite{hoang2019survey,wang2018deep,zhao2019deep}. Deep learning has a strong ability to extract features and it's easy and effective to establish an end-to-end fault diagnosis system \cite{khan2018review}. Deep learning methods such as Convolutional Neural Network (CNN) \cite{li2019understanding,abdeljaber20181,guo2018deep,li2020fault}, Recurrent Neural Network (RNN) \cite{lei2019fault,liu2018fault}, Auto-Encoder (AE) \cite{shao2018novel,yu2019selective} and Capsule Network (CN)\cite{chen2019deep,zhu2019convolutional} has proven effective on multiple problems. However, though deep learning system is powerful and easy to build, designing neural network architecture needs rich professional knowledge and debugging experience. To obtain an optimal architecture, a lot of experiments are needed, which leads to the time-consuming development of deep learning systems. What we want is an automated machine learning (AutoML) system that can automatically design neural network and adjust hype-parameters. 

\begin{figure}
	\centering
	\includegraphics[scale=0.66]{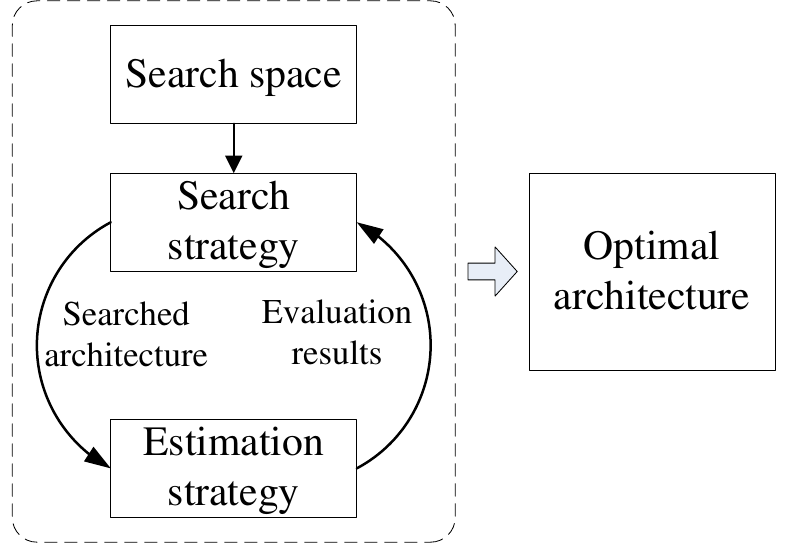}
	\caption{General process of neural architecture search.}
	\label{fig:process_of_nas}
\end{figure}

Fortunately, as a branch of AutoML, neural architecture search (NAS) is developing rapidly, and has become a new direction for deep learning\cite{elshawi2019automated,elsken2018neural,zoller2019survey}. The general process of the NAS is shown in Figure~\ref{fig:process_of_nas}. Given a specific learning task and a search space, NAS can automatically search for the optimal neural architecture. Generally speaking, a search strategy selects an architecture form predefined search space, then the selected architecture is evaluated by an estimation strategy. Next, the search strategy is updated according to the evaluation results. Repeat the above process, and finally get an optimal network architecture \cite{elsken2018neural}. 

In this paper, we propose a NAS method for fault diagnosis using reinforcement learning. A RNN is used as an agent (controller) to generate architectures that are trained with training dataset. And these architectures are evaluated with validation dataset to get accuracy. The accuracy is seen as a reward to controller, and the parameters of controller are updated using strategy gradient algorithm. We also utilize parameters sharing trick to accelerate the search process. The proposed method is proved to be effective on PHM 2009 Data Challenge gearbox dataset. Our contributions can be summarized into following aspects.

\begin{itemize}[leftmargin=*]
	\item We applied NAS in fault diagnosis for the first time. A reinforcement learning based NAS framework is developed to search for the optimal architecture. 
	\item We put forward several problems and challenges in the application of NAS and AutoML in fault diagnosis, and point out several directions for future research.
\end{itemize}

%%%%%%%%%%%%%%%%%%%%%%%%%%%%%%%%%%%%%%%%%%%%%%%%%%%%%%%%%%%%%%%%%%%%%%%%%%%%%%

\section{Related Work} 
\textbf{Fault diagnosis using deep learning}. Deep learning has been widely applied in fault diagnosis\cite{abdeljaber20181,guo2018deep,lei2019fault,li2019understanding,li2020fault,shao2018novel,yu2019selective}, and recently some novel network structures are proposed. \cite{zhu2019convolutional} proposes a novel capsule network with an Inception block for fault diagnosis. First signals are transformed into a time-frequency graph, and two convolution layers are applied to preliminarily extract features. Then an inception block is applied to improvethe nonlinearity of the capsule. After dynamic routing, the lengths of the capsules are used to classify the fault category. In order to obtain diversity resolution expressions of signals in frequency domain, \cite{huang2019improved} proposes a new CNN structure named multi-scale cascade convolutional neural network (MC-CNN). MC-CNN uses the filters with different scales to extract more useful information. To solve the problem that proper selection of features requires expertise knowledge and is time-consuming, \cite{pan2017liftingnet} proposes a novel network named LiftingNet to learn features adaptively without prior knowledge. LiftingNet introduced split layer, predict layer and update layer. And different kernel sizes are applied to improve learning ability.

\textbf{Neural architecture search.} The first influential job of NAS is \cite{zoph2016neural}. In this paper, author uses a RNN to generate the descriptions of neural networks, and train the RNN with reinforcement learning to maximize their excepted accuracy on validation dataset. The proposed method not only generate CNN, but also generate Long Short-Term Memory network (LSTM) cell. \cite{pham2018efficient} proposes a fast and inexpensive method named Efficient Neural Architecture Search (ENAS). This approach uses sharing parameters among child models to greatly reduce search time than above standard NAS. \cite{brock2017smash} employs an auxiliary HyperNet to generates the weights of a main model with variable architectures. And a flexible scheme based on memory read-writes is developed to define a diverse range of architectures. Unlike above approaches searching on a discrete and non-differentiable search space, \cite{liu2018darts} proposes a differentiable architecture search method named DARTS. This approach uses gradient descent to search architectures by relaxing the search space to be continuous. 

%%%%%%%%%%%%%%%%%%%%%%%%%%%%%%%%%%%%%%%%%%%%%%%%%%%%%%%%%%%%%%%%%%%%%%%%%%%%%%

\section{Methods}
According to \cite{zoph2016neural}, neural network can be typically specified by a variable-length string, so it can be generated by RNN. In this section, we will use a RNN as controller to generate a CNN with reinforcement learning. Given a search space, CNN can be designed by RNN, and RNN is trained with a policy gradient method to maximize the expected accuracy of the generated architectures.

\subsection{Search Space}
Our method searches for the optimal convolution kernel combination in a fixed network structure. Several typical network structure can be selected such as Inception structure, ResNet structure, DenseNet structure and so on. In this paper, we search the optimal architecture in a ResNet structure which is shown in Figure~\ref{fig:nas-example}. The inputs are first feed into a fixed stem layer, and then followed by several residual blocks, where convolutional kernel of each layer is generated by RNN. Finally, global average pooling layer flattens the features maps and classifier outputs the classification probability. There are two layers and a skip connection in a residual block. Here we use six different convolutional kernels: 

\begin{itemize}[leftmargin=*]
\item $3 \times 1$ kernel with dilation rate $d=1$
\item $3 \times 1$ kernel with dilation rate $d=2$
\item $3 \times 1$ kernel with dilation rate $d=3$
\item $5 \times 1$ kernel with dilation rate $d=1$
\item $5 \times 1$ kernel with dilation rate $d=2$
\item $5 \times 1$ kernel with dilation rate $d=3$
\end{itemize}

\begin{figure}
	\centering
	\includegraphics[scale=0.66]{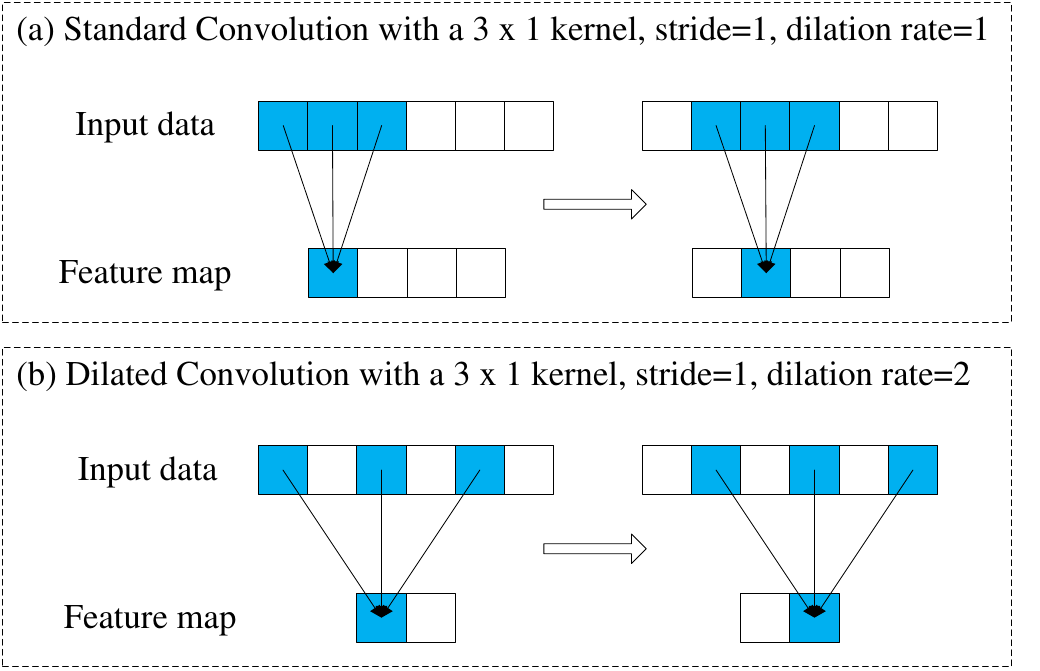}
	\caption{An example dilated convolution compared with standard convolution.}
	\label{fig:dilated}
\end{figure}

Dilated convolution is to inject holes into the standard convolution kernel to increase the receptive field \cite{yu2015multi}. Compared with the standard convolution operation, the dilated convolution has one more hyperparameter called dilation rate $d$, which refers to the number of kernel intervals. An example of dilated convolution compared with standard convolution is shown in Figure~\ref{fig:dilated}. 

In this paper, we set 4 blocks in the ResNet structure, each layer has 6 different convolution kernels to choice, so there are $6^{(4 \times 2)} \approx 1.68 \times 10^{6}$ possible architectures. Our aim is to search the optimal architecture in such a large search space.

\subsection{Designing CNN using Recurrent Neural Network}

\begin{figure}
\centering
\includegraphics[scale=0.64]{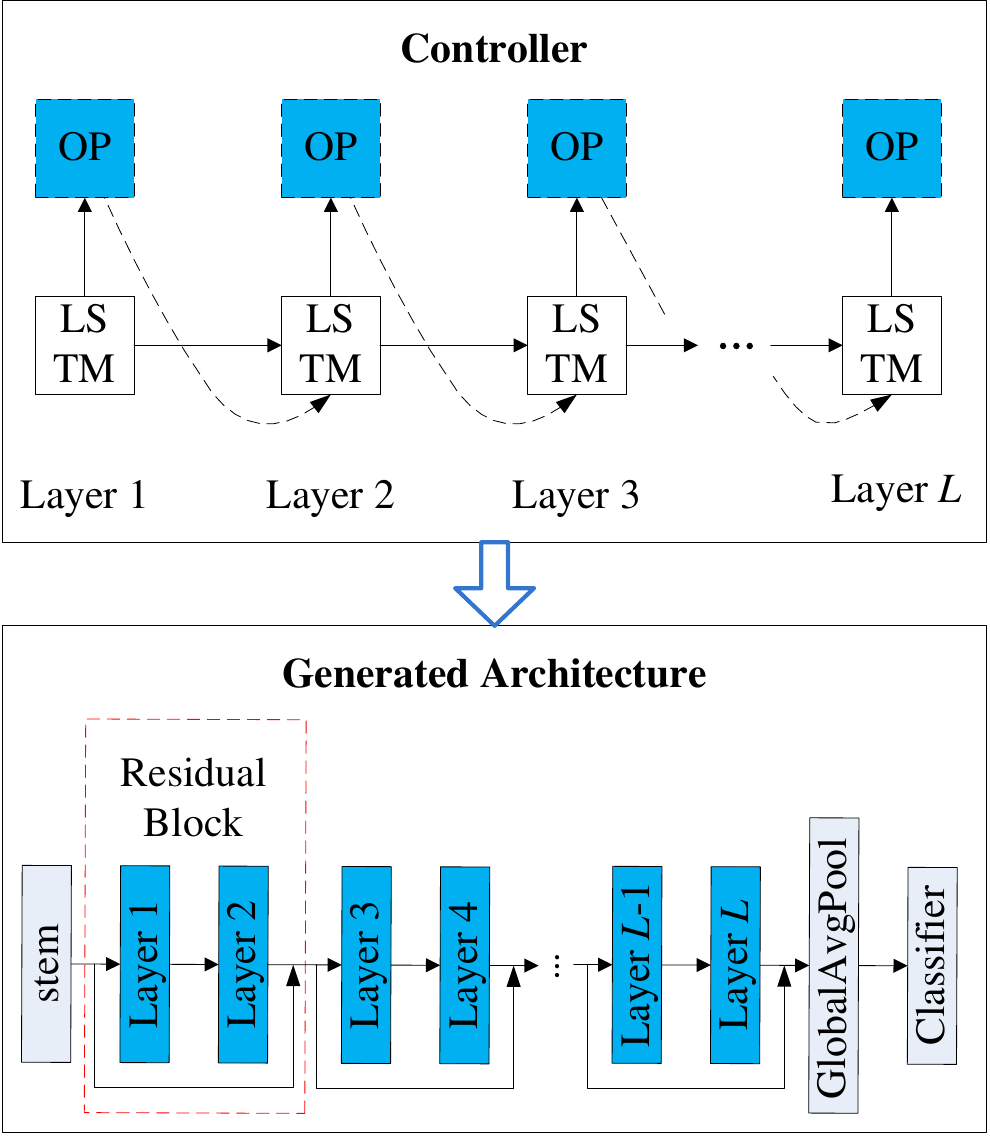}
\caption{An example run of designing CNN using RNN controller.}
\label{fig:nas-example}
\end{figure}

Since a neural network can be encoded by a variable-length string, it's possible to use RNN, a controller to generate such string. Here, six different convolution kernels are encoded as Numbers $0 \sim 5$, so different combinations of Numbers represent different network architectures. In this paper, we use LSTM to generate such Numbers combinations, as shown in Figure~\ref{fig:nas-example}. For LSTM, the output probability distribution of six convolution kernels is obtained by \textit{softmax}, and a certain kernel is sampled form such distribution. For example, for the first layer of CNN, the controller outputs a softmax probability distribution: $[0.1, 0.1, 0.2, 0.3, 0.2, 0.1]$. And the probability of the fourth convolution kernel being sampled is 0.3, and it is most likely to be sampled. Then this sampled convolution kernel is the convolution operation of the first CNN layer. Next, the embedding of sampled Number is used as input to the LSTM to generate the convolution kernel of the next layer. And so on, until the convolution kernels of all layers are generated.

\subsection{Training With Reinforcement Learning}
In reinforcement learning, there are two main parts: \textit{agent} and \textit{environment}. Agent gets rewards by interacting with the environment to learn the corresponding strategies. In reinforcement learning based NAS, agent is the RNN controller, environment is the search space, the validation accuracy of sampled model is reward. The generated CNN architecture by controller is trained using training dataset $D_{train}$, and this CNN is evaluated using validation dataset $D_{val}$ to get reward $R$. Then controller is updated using the reward. To find optimal architecture, we need to maximize the expected reward of controller: 

\begin{equation}
	J(\theta_{c}) = \mathbb{E}_{P(a_{1:L}; \theta_{c})}[R]
\end{equation}

Where $\theta_{c}$ is the parameters of controller, $a_{1:L}$ is a list of convolution kernels sampled by controller to generate a CNN, $P(a_{1:L}; \theta_{c})$ is the probability that $a_{1:L}$ is sampled. But the reward signal is not differentiable, we use the policy gradient algorithm to iteratively update $J$ \cite{williams1992simple}:

\begin{equation}
	\begin{aligned}
		\nabla \theta_{c} J\left(\theta_{c}\right) &=\sum_{l=1}^{L} E_{P\left(a_{1: L} ; \theta_{c}\right)} G_{\theta_{c}} \\
		G_{\theta_{c}} &= \nabla \theta_{c} \log P\left(a_{l} | a_{(l-1): 1} ; \theta_{c}\right) R
	\end{aligned}
\end{equation}

An empirical approximation of the above quantity is:

\begin{equation}
\frac{1}{m} \sum_{k=1}^{m} \sum_{l=1}^{L} \nabla_{\theta_{c}} \log P\left(a_{l} | a_{(l-1): 1} ; \theta_{c}\right) R_{k}
\end{equation}

Where $m$ is the number of different architectures that the controller samples in one batch, $L$ is the number of convolution kernels our controller has to predict in each CNN, and $R_{k}$ is the reward of $k$-th sampled architecture. Above updating rule is an unbiased estimate and has a very high variance. In order to reduce the variance of this estimate, we use a baseline function to this updating rule \cite{zoph2016neural}:

\begin{equation}
\frac{1}{m} \sum_{k=1}^{m} \sum_{l=1}^{L} \nabla \theta_{c} \log P\left(a_{l} | a_{(l-1): 1} ; \theta_{c}\right)\left(R_{k}-b\right)
\end{equation}

Where $b$ is an exponential moving average of the previous architecture validation accuracies.

\subsection{Accelerate Training using Parameters Sharing}

As we all know, training a neural network from scratch is time-consuming. In the process of search, a sampled architecture need to be trained from scratch to obtain it's reward. This can be very time-consuming and inefficient when the number of search epochs is particularly large. To reduce the cost of searching, the weight sharing mechanism is applied in training process. We don't train the sampled architecture form scratch, but train the model with only one mini-batch data, and the trained convolution kernels will be reused in next search epoch \cite{pham2018efficient}. There are many repeated convolution operations among architectures, and weight sharing can prevent them from being repeatedly trained. This greatly improves the efficiency of search process.

\subsection{Neural Architecture Search Pipeline}

In each search epoch, RNN will generate a number of architectures according to the output probability distribution. These architectures will be trained with signal mini-batch training data, and their rewards are obtained using validation data. Then the controller is update according to Eq.~(4). Above search process is then repeated until the maximum number of search epochs is reached. Finally, $M$ architectures are generated by the trained controller, and the architecture with the highest validation accuracy is selected as the final architecture, and trained from scratch. The whole process of neural architecture search for fault diagnosis is shown is Algorithm~\ref{alg:nas} and Algorithm~\ref{alg:sample}.

\begin{algorithm}[tb]
	\caption{NAS for fault diagnosis.}
	\label{alg:nas}
	\begin{algorithmic}
		\STATE {\bfseries Input:} Search space $S(n, L)$ with $n$ choices per layer and $L$ layers in total, search epochs $N$, controller train steps $N_{c}$ in each search epoch, number of sampled architectures $m$ and $M$, parameters of candidate convolution kernels $\theta_{kernel}(n, L)$, parameters of controller $\theta_{c}$, training data $D_{train}$, validation data $D_{val}$.
		\STATE {\bfseries Initialize} $\theta_{kernel}(n, L)$ and $\theta_{c}$.
		\FOR{$i=1$ {\bfseries to} $N$}
			\FOR{$x, y$ {\bfseries in} $D_{train}$}
				\STATE Sample an architecture $A$ with convolution kernels $\theta_{kernel}(A)$, train it using $x, y$.
				\STATE Save the trained convolution kernels $\theta_{kernel}(A)$.
			\ENDFOR
			\FOR{$j$ {\bfseries to} $N_{c}$}
				\STATE Sample $m$ architectures, their rewards $R_{1:m}$ are obtained by $D_{val}$.
				\STATE Update controller according to Eq.~(4).
			\ENDFOR
		\ENDFOR
		\STATE Sample $M$ architectures, get their rewards $R_{1:M}$. Find the architecture $A$ with highest reward $R_{max}$, train it from scratch.
	\end{algorithmic}
\end{algorithm}

\begin{algorithm}[tb]
	\caption{Sample architecture using LSTM.}
	\label{alg:sample}
	\begin{algorithmic}
		\STATE {\bfseries Input:} Input size of LSTM $I$, hidden size of LSTM $H$, number of LSTM layers $L$, number of convolution kernels in each CNN layer $N$.
		\STATE {\bfseries Initialize} controller $LSTM(I, H, L, N)$. 
		\STATE Get the embedding of first convolution kernel in CNN $w_{0}$
		\FOR{$i=1$ {\bfseries to} $N$}
			\STATE Use embedding $w_{i-1}$ as the input of $LSTM$, get the output probability distribution $P_{i}$.
			\STATE Get the number of a convolution kernel $a_{i}$ based on probability sampling $P_{i}$.
			\STATE Get the embedding $w_{i}$ of sampled number $a_{i}$.
		\ENDFOR
		\STATE Obtain the sampled architecture $A=\{a_{1}, a_{2}, \cdots, a_{N}\}$.
	\end{algorithmic}
\end{algorithm}

%%%%%%%%%%%%%%%%%%%%%%%%%%%%%%%%%%%%%%%%%%%%%%%%%%%%%%%%%%%%%%%%%%%%%%%%%%%%%%

\section{Experiments}

\subsection{PHM-2009 Dataset}
\begin{figure}
	\centering
	\includegraphics[scale=0.5]{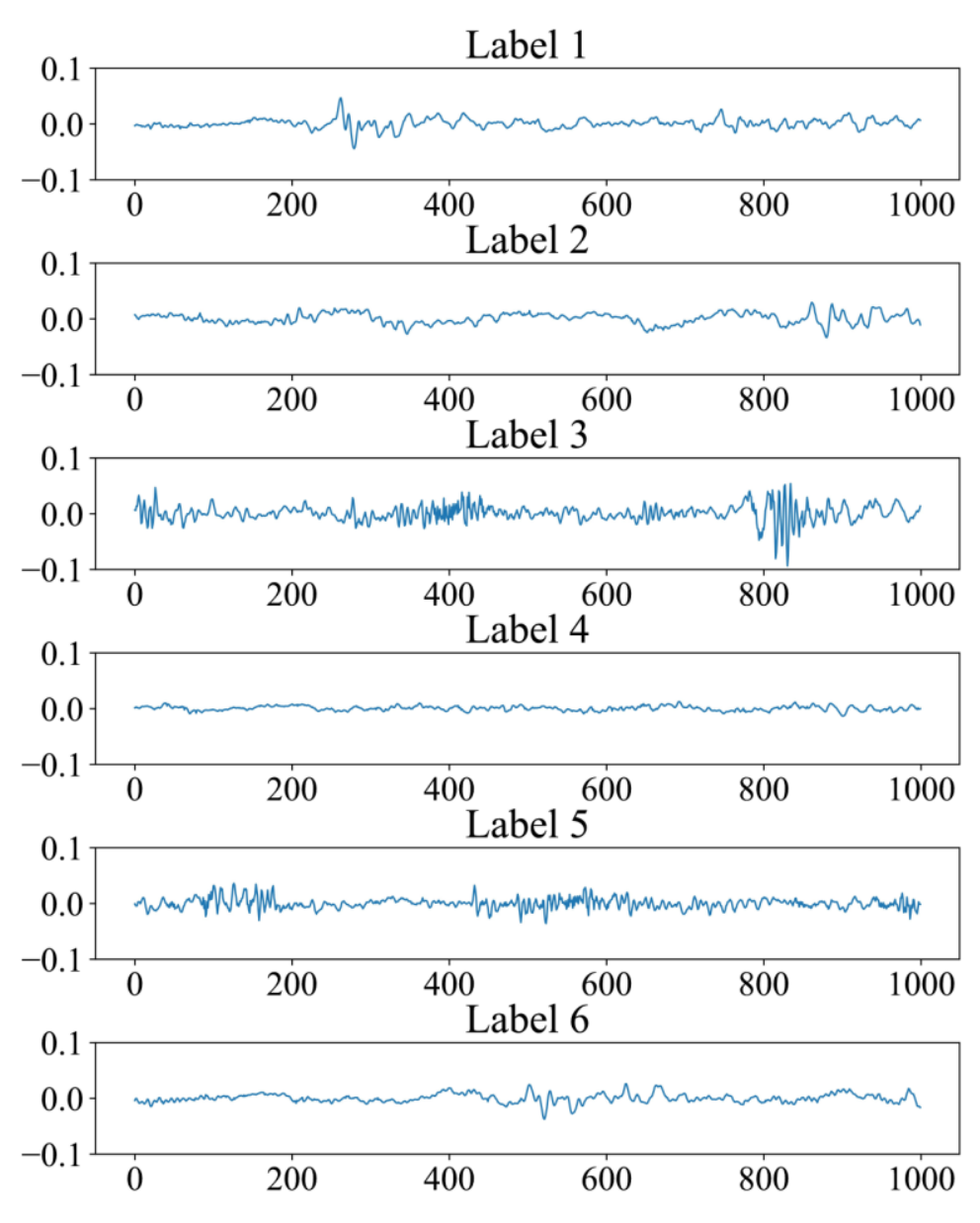}
	\caption{An vibration signals example of six labels.}
	\label{fig:signals-example}
\end{figure}

In this paper, we use gearbox dataset of PHM-2009 Data Challenge to study NAS for fault diagnosis. This dataset is a typical industrial gearbox data which contains 3 shafts, 4 gears and 6 bearings. Two sets of gears, spur gears and helical gears are tested. There are six labels in this dataset. For each label, there are 5 kinds of shaft speed, 30, 35, 40, 45 and 50 Hz, and two modes of loading, high and low loading. We do not distinguish between these working conditions under each label. The raw vibration signals of this dataset are very long, so we use a sliding window with a length of 1000 and a step length of 50 to segment the signals to obtain training, validation and testing samples. These signals are normalized to $[-1, 1]$. Finally, we obtain 22967 training samples, 2552 samples and 6380 samples. An vibration signals example of six labels is shown Figure~\ref{fig:signals-example}.

\subsection{Training Details}
For the controller, we set the input size $I$ and hidden size $H$ of LSTM to be 64, number of layer to be 1. We use Stochastic Gradient Descent (SGD) with learning rate of 0.01 to train the controller. In each search epoch, we train the controller for $N_{c}=5$ steps. In each train step, we sample $m=20$ architectures. For the sampled architecture training, we use Adam with learning rate of 0.001 and L2 regularization of $1e-4$. For the ResNet structure, we set 4 residual blocks and each block contains 2 layers. Each layer is followed by a down-sampling layer with a convolution kernel of $3 \times 1$ and s step size of 2. The number of channels in the first block is 8, then doubles as it passes through the down-sampling layer. We set search epochs $N=200$, batch size to be 128. After the search, $M=100$ architectures are sampled to be evaluated, and the architecture with highest validation accuracy is found as the final model. The final model will be trained from scratch using Adam with learning rate of 0.001 and batch size of 128. The final result of searched model is evaluated on testing dataset. In the above process, learning rate is adjusted using CosineAnnealingLR. The code is implemented using PyTorch 1.3, using a signal Tesla K80 GPU. The whole search time is 1.6 hours.

\subsection{Results}
Table~\ref{tab:result} summarizes the results of NAS and six manually designed models. All layers of M1 model are the first convolution kernel in the search space, M2 is the second, and so on. Note that our method is searching in the ResNet structure, so all compared models are variants of ResNet. We can also search the architectures in Inception structure, DenseNet structure and so on. 

\begin{figure}[ht]
	\centering
	\includegraphics[scale=0.64]{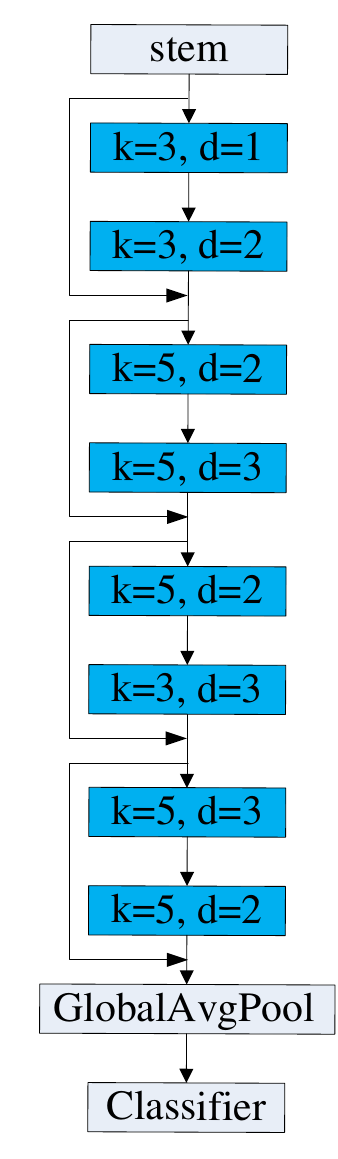}
	\caption{The architecture of searched model.}
	\label{fig:searched-model}
\end{figure}

\begin{figure}
	\centering
	\includegraphics[scale=0.5]{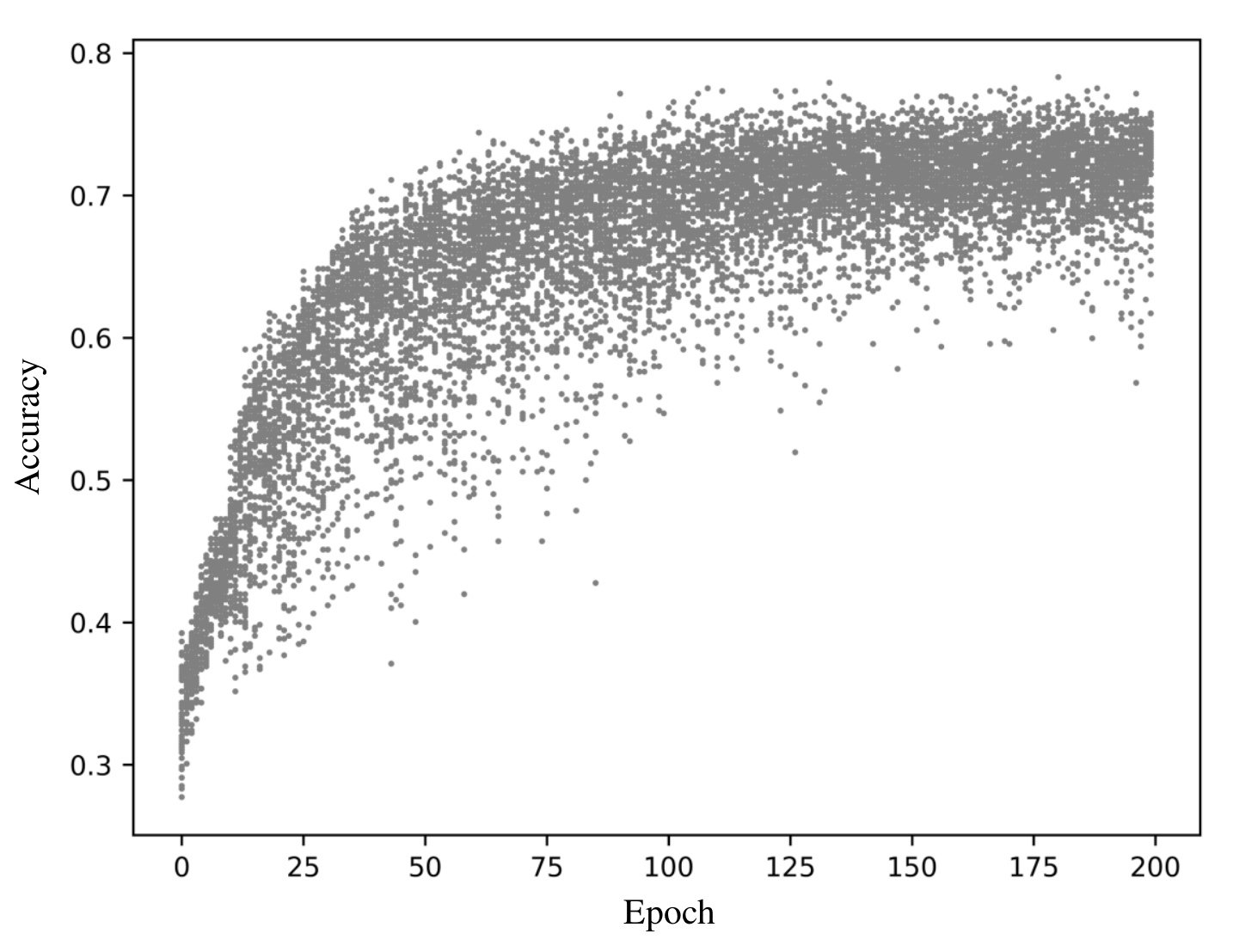}
	\caption{Validation accuracy of training process, and 50 models are sampled in each search epoch.}
	\label{fig:acc}
\end{figure}

The searched architecture is shown in Figure~\ref{fig:searched-model}, and it achieved accuracy of 78.91\%. Six manually designed models achieved at most accuracy of 76.22\%. This indicates that the method of NAS based on reinforcement learning is effective, and the controller gets rewards through the sampled models and constantly update the parameters in the direction of obtaining more excellent models. In addition, after each search epoch, 50 architectures were sampled to get their validation accuracy. Figure~\ref{fig:acc} shows the trends of those accuracy rates throughout the search process. We can see that accuracies increase gradually. It indicates that the repeated use of convolution kernels is effective, and it does improve the performance and stability of the entire ResNet structure to accurately evaluate the performance of sampled architectures.

\begin{table}
\tbl{Testing accuracy comparasion of searched model and hand-designed model.}
{\tabcolsep14pt
\begin{tabular}{@{}lcl@{}}\toprule 
Model & Architecture & Accuracy (\%)\\\colrule
M1 & k=3, d=1 & 68.43\\
M2 & k=5, d=1 & 71.19\\
M3 & k=3, d=2 & 69.37\\
M4 & k=5, d=2 & 72.46\\
M5 & k=3, d=3 & 72.77\\
M6 & k=5, d=3 & 76.22\\
Searched & / & \textbf{78.91} \\\botrule
\end{tabular}}
\label{tab:result}
\end{table}

%%%%%%%%%%%%%%%%%%%%%%%%%%%%%%%%%%%%%%%%%%%%%%%%%%%%%%%%%%%%%%%%%%%%%%%%%%%%%%
\section{Discussions}

In this paper, we have initially shown the application of NAS in fault diagnosis and proved its effectiveness. However, the application of NAS in fault diagnosis has just started, and there are still many challenges to realize the automatic design of deep learning models for fault diagnosis. We have summarized the following problems to be solved:

\begin{itemize}[leftmargin=*]
\item In this paper, we just search the optimal architecture in ResNet structure, which has great limitations. We are more interested in how to automatically design more novel and complex structures, not limited by the existing structures or the number of layers, and the searched models have better performance. It is currently the most challenging problem. 

\item Reinforcement learning based NAS is also a proxy NAS that will cost a lot of time. In this paper, due to the small dataset, the small network size, and the small number of training epochs, the entire search took only 1.6 hours, of which the time of training controller accounts for a large part. How to use more effective search methods for fault diagnosis is the problem that needs to be solved urgently.

\item In this paper, we only evaluate the testing accuracy of sampled architectures, but did not focus on the amount of parameters of the searched architectures (which determines the storage space occupied by the model) and the amount of calculation (which determines the speed of the model). When the model is deployed in an embedded terminal, the amount of parameters and calculations become very important. How to search for a model with a small number of parameters and a small amount of calculation but with high accuracy is also a difficult problem for future research.

\item Not only limited to neural architecture search, realizing the automation of machine learning in fault diagnosis is a wider and more difficult problem. The data of industrial equipment is huge and complicated, the preprocessing of data is difficult, and the working conditions are changing. From data collection to data pre-processing, to feature engineering and modeling, to model testing and tuning parameters, the entire development process cycle is long and time consuming. Automating machine learning in the field of PHM is a more difficult challenge.

\end{itemize}

%%%%%%%%%%%%%%%%%%%%%%%%%%%%%%%%%%%%%%%%%%%%%%%%%%%%%%%%%%%%%%%%%%%%%%%%%%%%%%
\section{Conclusions}

In this paper, we develop a method of neural architecture search for fault diagnosis. It's the first time that NAS technology has been used to automatically generate deep learning model for fault diagnosis. We use RNN as a controller to generate architectures in ResNet search space, and train the controller with reinforcement learning. Results show that NAS is effective to find a model with better performance than manually designed models.

\begin{acknowledgement}
The work of Yang Hu is supported by the National Nature Science Foundation of China, grant number 61703431. The work of Mingtao Li was supported by the Youth Innovation Promotion Association CAS. The computing platform is provided by the STARNET cloud platform of National Space Science Center Public Technology Service Center.
\end{acknowledgement}

%\section*{Appendix A. The Relation Between {\it R} and {\it V}}
%
%Appendices should be used only when absolutely necessary.  They should come before the References.  If there is more than one appendix, label them alphabetically.  Number the displayed equations occurring in the Appendix in this way, e.g.,~(A.1), (A.2), etc
%\def\theequation{A.1}
%\begin{eqnarray}
%\frac{S_R (f)}{R^2} \cong \frac{\alpha _H }{nV_{\mbox{eff}} }\frac{1}{f^k}.
%\end{eqnarray}
%\begin{eqnarray}
%\mu (n,t) = \frac{{\sum}^{\infty}_{i=1}(d_i < t, N(d_i)=n)}{{\int}^{t}_{\sigma=0}1(N({\sigma})=n)d{\sigma}}
%\end{eqnarray}

%\section*{CItations and References}
%
%Type references in continuous format in 9pt Times New Roman with single line spacing.   The references need to be in a proper style, i.e., the
%Chicago bibliography style. Examples: \cite{lamport94,knuth79}. For more information, please visit: https://ctan.org/pkg/
%% We should include an example bibfile with a couple of entries as well as citation in the body text
\bibliography{sample}
\bibliographystyle{Chicago}

\end{document}